\documentclass[runningheads]{llncs}
\usepackage{pdfpages}
\usepackage[width=122mm,left=12mm,paperwidth=146mm,height=193mm,top=12mm,paperheight=217mm]{geometry}

\usepackage[colorlinks=true,bookmarks=false]{hyperref}

\newcommand{\app}{\raise.17ex\hbox{$\scriptstyle\sim$}}

\newlength\savewidth
\usepackage{graphicx}
\usepackage{comment}
\usepackage{amsmath,amssymb} 
\usepackage{color}
\usepackage{multirow}
\usepackage{bm}
\usepackage[caption=false]{subfig}
\usepackage[colorlinks=true,bookmarks=false]{hyperref}
\DeclareMathOperator{\softmax}{softmax}
\DeclareMathOperator*{\argmax}{arg\,max}

\usepackage{float}
\newcommand{\minisection}[1]{\vspace{0.04in} \noindent {\bf #1}}

\begin{document}

\pagestyle{headings}
\mainmatter

\title{Simple and effective localized attribute representations for zero-shot learning} 

\author{Shiqi Yang, Kai Wang, Luis Herranz, Joost van de Weijer\\
\email{{syang,kwang,lherranz,joost}@cvc.uab.es}}
\institute{Computer Vision Center, Autonomous University of Barcelona}

\maketitle

\begin{abstract}

Zero-shot learning (ZSL) aims to discriminate images from unseen classes by exploiting relations to seen classes via their semantic descriptions. Some recent papers have shown the importance of localized features together with fine-tuning the feature extractor to obtain discriminative and transferable features. However, these methods require complex attention or part detection modules to perform explicit localization in the visual space. In contrast, in this paper we propose localizing representations in the semantic/attribute space, with a simple but effective pipeline where localization is implicit. Focusing on attribute representations, we show that our method obtains state-of-the-art performance on CUB and SUN datasets, and also achieves competitive results on AWA2 dataset, outperforming generally more complex methods with explicit localization in the visual space. Our method can be implemented easily, which can be used as a new baseline for zero shot-learning. In addition, our localized representations are highly interpretable as attribute-specific heatmaps.
\end{abstract}

\section{Introduction}

Visual classification with deep convolutional neural networks has achieved remarkable success~\cite{he2016deep,simonyan2014very}, even surpassing humans in some benchmarks~\cite{he2015delving}. This success, however, requires that the training data contain enough images per class (tens or hundreds of images), which is often not the case in practice, and visual data to learn new classes may be scarce (i.e. few-shot learning -FSL-) or inexistent (i.e. zero-shot learning -ZSL-). Humans, in contrast, are able to infer new classes from few or even no visual examples, just from a semantic description that connects them to known concepts (e.g. a zebra is like a horse but with stripes). Thus, ZSL is a desirable capability in computer vision systems, allowing them to recognize a much larger set of classes via their semantic descriptions.

The key component of a ZSL system is the \textit{semantic model}, which connects \textit{seen} and \textit{unseen} classes in a common semantic space that enables the transference of seen visual representations to infer unseen  classes. The most common semantic spaces are visual attributes, word embeddings and textual descriptions. We focus on visual attributes. In addition, generalized zero-shot learning (GZSL) addresses the setting where the test image could belong to seen classes (in addition to unseen classes). In this case the main challenge is the inherent bias towards seen classes. Thus, discriminative and transferable representations, together with properly designed semantic spaces, are key for effective and unbiased inference on unseen classes. In this paper we focus on attributes as semantic model and learning representations that are transferable to unseen classes with low bias.

The common approach is to align visual and semantic representations in a common embedding space via a ranking loss or metric learning losses. The visual representation is extracted with a visual model and the semantic representation is a mapping of the class to the semantic space (e.g. a class prototype in terms of attributes). During inference, seen and unseen classes are mapped to the common embedding space and the class nearest to the visual representation is selected.

\begin{figure}[tb]
	\centering
	\subfloat[Global representation\label{fig:global_rep}]{
		\includegraphics[scale=0.55]{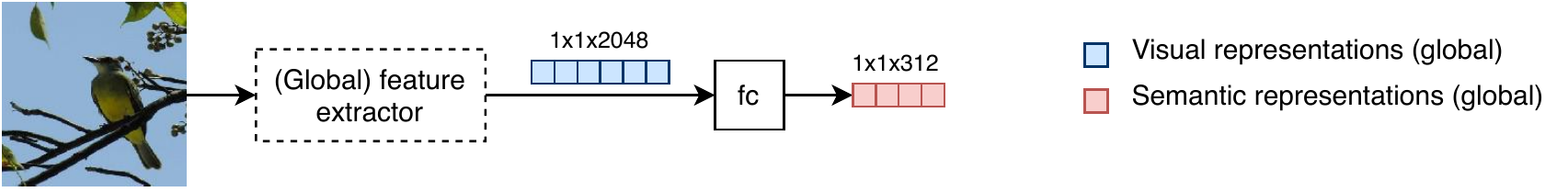}
	}\\
	\subfloat[Part-based representation\label{fig:parts}]{
		\includegraphics[scale=0.55]{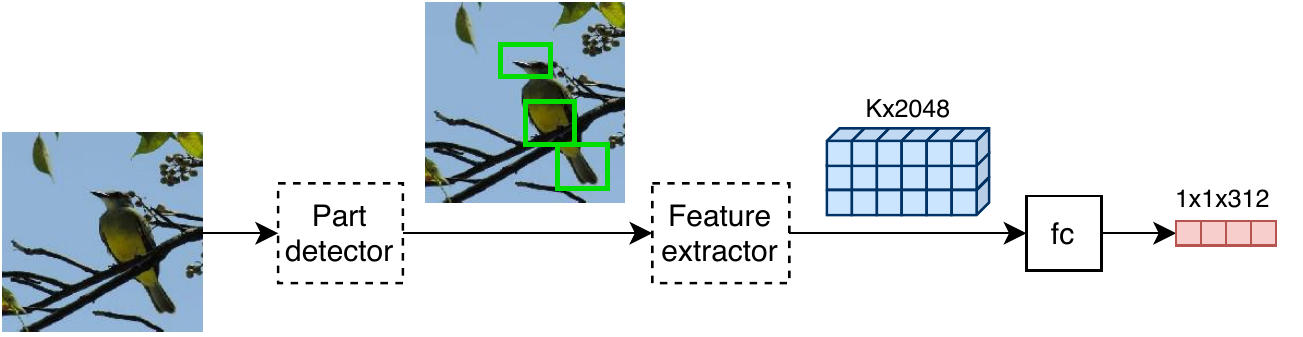}
	}\\
	\subfloat[Attention-based representation\label{fig:attention}]{
		\includegraphics[scale=0.55]{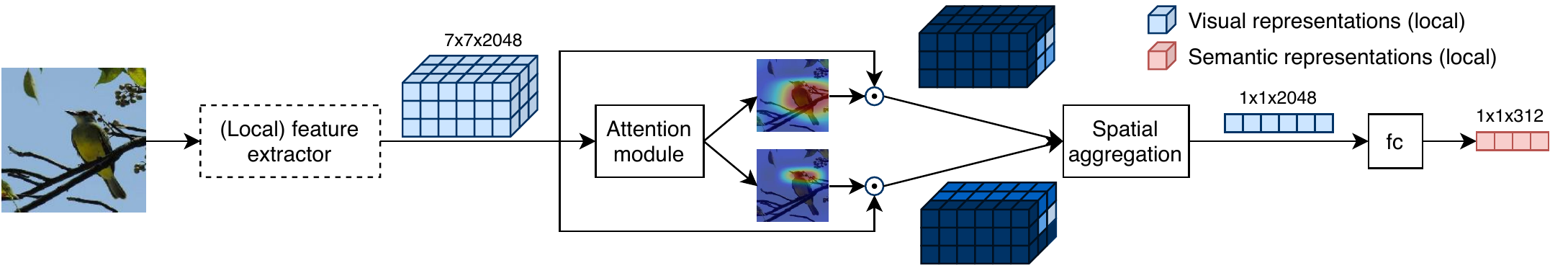}
	}\\
	\subfloat[Proposed representation: localized attributes\label{fig:proposed}]{
		\includegraphics[scale=0.55]{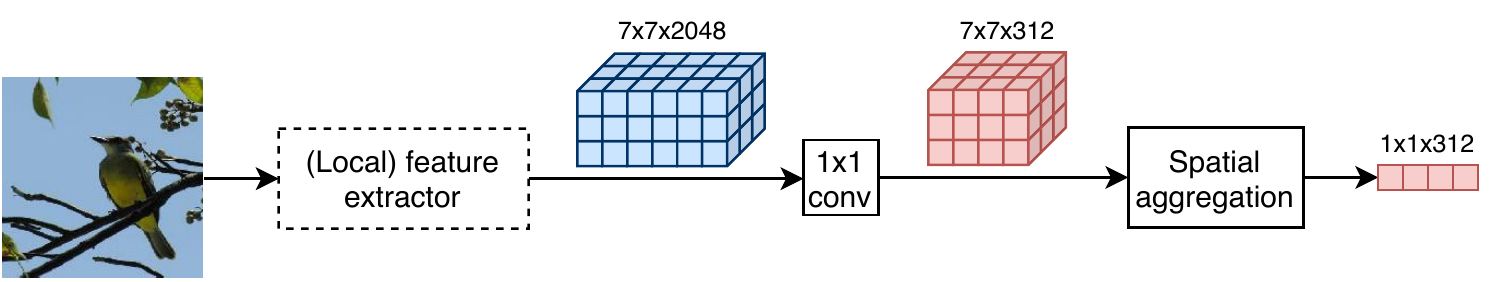}
	}
	\caption{Semantic representations for (G)ZSL. We propose (c) localized attribute representations that explicitly localize and disentangles attribute information, in contrast to (a) global representations, and (b) part-based and (c) attention-based approaches that focus on few regions in the visual space and do not disentangle attribute information until the global representation.}
	\label{fig:motivation}
\end{figure}

The most common representations in ZSL are global visual features extracted from an (ImageNet-)pretrained feature extractor (see Fig.~\ref{fig:architechture_global_visual}),  which are even readily available off-the-shelf from previous works~\cite{Xian2018ZSLGBU,zhang2017learning}. These global visual features are then projected to the semantic space~\cite{Akata2016LabelEmbedding,Frome2013Devise} or to an intermediate space~\cite{zhang2015zero}, where the comparison with semantic representations takes place. Most papers have focused on designing and learning a good visual-semantic alignment. Notably, Zhang \textit{et al.}~\cite{zhang2017learning} suggest that the choice of the embedding space is crucial, and argue that the projection to low dimensional semantic spaces or intermediate spaces shrinks the variance of the visual feature, limiting its discriminability and increasing the so-called hubness problem~\cite{radovanovic2010hubs}. They suggest that the visual space is more discriminative and robust to hubness, and propose to embed classes directly in the visual space and then perform nearest neighbor search. In this paper we argue that this conclusion only holds for global representations, and that the choice of space where features are localized is even more critical. 

Little attention has been paid to the role of locality in the design of good semantic representations that are discriminative and transferable to unseen classes.
This is particularly critical in fine-grained scenarios where the differences between classes are highly local and subtle (e.g. the color of the beak or the tail could be the only aspect that can discriminate between two classes of birds). This has been confirmed in recent analysis~\cite{sylvain2019locality}. In this paper we show that the semantic space is indeed a better choice to project local features, and using a suitable spatial aggregation strategy, the resulting global semantic feature remains highly discriminative and effective for classification of unseen (and seen) classes.

In this paper we focus on localized semantic representations, and, in particular we propose \textit{localized attributes} as representations. They can be obtained easily if we rethink the ZSL pipeline and switch the order between spatial aggregation (i.e. local to global) and projection to the semantic space (see Fig.~\ref{fig:proposed}). We also investigate how a proper choice of the spatial aggregation mechanism can significantly boost the performance of the semantic representation in some datasets. We show that a simple convolutional layer and global max pooling (GMP) are enough to achieve highly competitive performance and outperform most part-based and attention-based methods. Its simplicity also entails multiple advantages, including  easier and efficient training (simply using standard cross-entropy loss), no additional hyperparameters and better transferability to unseen classes (i.e. seen and unseen accuracies are more balanced).

Localized representations have been proposed earlier in ZSL~\cite{elhoseiny2019creativity,Xie_2019_CVPR,zhu2018generative,zhu2019semantic}. However, they are mostly limited to the detection or discovery of discriminative parts~\cite{elhoseiny2019creativity,zhu2018generative} (see Fig.~\ref{fig:parts}) or attention mechanisms in the visual space~\cite{Xie_2019_CVPR,zhu2019semantic} (see Fig.~\ref{fig:attention}), rather than explicitly localizing attributes in a local semantic space as we propose here. In addition, part detectors are often trained separately with additional part-specific annotations (e.g. bounding boxes). Sometimes, extracting local features may also require larger images~\cite{zhu2019semantic}. The number of extracted regions or attention maps is typically low (typically 2-15), in contrast to our localized attributes (85-312 in typical datasets). In addition, the few features extracted from detected parts or attention maps are not necessarily disentangling attribute information, while we obtain one dedicated map for every attribute.

In short, we summarize our contributions as follows:
\begin{itemize}
	\item We propose a simple and effective localized attribute representation (SELAR) for (G)ZSL, which is both discriminative and transferable to unseen classes. The representation is also  interpretable as attribute-specific heatmaps. 
	\item We study the role of the aggregation mechanism to improve localization and reduce the bias. This analysis shows that global max pooling in the localized attribute space leads to significant performance gains, especially improving performance on the unseen classes. 
	\item We achieve state-of-the-art performance on the SUN and CUB dataset. Notably, our method, which implicitly localizes the attributes, outperforms other more complicated methods with networks with explicit localization, such as attention-based~\cite{Xie_2019_CVPR,zhu2019semantic} and part-based~\cite{elhoseiny2019creativity,zhu2018generative} methods.
\end{itemize}

\section{Related work}
\paragraph{Zero-shot learning} The original ZSL task focuses on achieving good predictions on unseen classes.  
Early approaches tackle this problem via visual-semantic  alignment in a common space~\cite{akata2015label,akata2015evaluation,ekodirov_cvpr2017,Frome2013Devise,Norouzi2014ConSe,romera2015embarrassingly,socher2013zero,xian2016latent,zhang2015zero}. The common space can be the semantic space, the visual space~\cite{li2019rethinking,zhang2017learning} or an intermediate space~\cite{zhang2015zero}. The alignment can be achieved via linear projections~\cite{akata2015label,akata2015evaluation,Frome2013Devise}, non-linear projections~\cite{socher2013zero,xian2016latent} or combinations of seen embeddings~\cite{Changpinyo2016ZSL,Norouzi2014ConSe}. Typically, a ranking loss is used to enforce alignment, but L2 loss~\cite{zhang2017learning} and adversarial loss~\cite{zhu2018generative} have also been used.

\paragraph{Generalized zero-shot learning}
This more challenging, yet more realistic, setting evaluates the classifier on the union of seen and unseen classes~\cite{chao2016empirical}. The additional problem of bias towards seen classes becomes critical for good GZSL performance, and requires specific techniques to address it~\cite{chao2016empirical,felix2018multi,li2019rethinking,liu2018generalized,XianCVPR2019a}. Several of these works relax some assumptions of the GZSL setting and achieve better performance. For example, one of such relaxations is considering that the descriptions of unseen classes are available during training. In that case, a generative model can be trained to generate synthetic features of unseen classes, and then combine them with real seen samples to train a joint and balanced classifier for both seen and unseen classes~\cite{felix2018multi,Mishra_2018_CVPR_Workshops,xianCVPR18,XianCVPR2019a}. Another assumption is having access to unseen images and labels which can help to calibrate the bias between the scores of seen and unseen classes~\cite{chao2016empirical,liu2018generalized}. In this paper, we assume that nor the unseen descriptions are available during training, nor we can calibrate the classifier. 

\paragraph{Localized features}
Most (G)ZSL approaches focus on the role of the classifier and the semantic models, directly relying on global representations extracted by a pretrained classifier (typically a ResNet-101 trained on ImageNet). The potential of local representations for (G)ZSL has been explored only recently in two directions: part detection~\cite{elhoseiny2019creativity,zhu2018generative}  and attention mechanisms~\cite{Xie_2019_CVPR,zhu2019semantic}. In the former group, Zhu \textit{et al.} uses a part detector trained for fine-grain recognition, where a fixed number of parts is extracted (e.g. seven parts for birds in CUB dataset, such as beak, belly, wings, etc.). Then adversarial learning is used to align semantic and visual representations. The model was improved including a loss encouraging creativity in the model~\cite{elhoseiny2019creativity}. The main limitation of these approaches is that the part detector requires additional and expensive annotation data (i.e. part ids, bounding boxes) to train the part detector. In the latter group, attention mechanisms focus on discovering discriminative regions. AREN~\cite{Xie_2019_CVPR} includes an attention layer, combined with an adaptive thresholding mechanism and a second order pooling representation. SGMA~\cite{zhu2019semantic} first computes part attention maps (only two maps in their case), which subsequently guide a part extractor where local features are extracted. In general, the feature extractor in these methods is fine-tuned to improve the localization ability. Part detectors and attention mechanisms are significantly more complex and arguably more difficult to train (having additional hyperparameters) than the proposed approach, and essentially different since the attributes are not explicitly localized but only a few visual regions.

\section{Zero-shot learning with localized attribute representations}
\subsection{Task Definition}

In the ZSL task, the training set contains seen classes and is defined as $\mathcal{S}\equiv \{(\mathbf{x}_i^s,y_i^s)\}_{i=1}^{N_s}$, where $i$ denotes the $i$-th image of the seen class and $y_i^s\in\mathcal{Y}^{\mathcal{S}}$ is its  class label. The test set contains unseen classes and is defined as $\mathcal{U}\equiv \{(\mathbf{x}_j^u,y_j^u)\}_{j=1}^{N_u}$. The sets of seen and unseen classes are disjoint, i.e.  $\mathcal{Y}^{\mathcal{S}}\cap\mathcal{Y}^{\mathcal{U}}=\emptyset$. The semantic information about a particular class $y$ is obtained by the class embedding function as $\psi\left(y_i\right)$. In the case of attribute-based representations with $L$ attributes, the class prototype $\psi_i=\psi\left(y_i\right)$ is simply a $L$-dimensional (binary or real valued) attribute vector encoding the presence or absence of each attribute. In this way, the semantic information about all seen classes can be conveniently captured in 
$\left|\mathcal{Y}^\mathcal{S}\right| \times L$-dimensional attribute matrix $A^\mathcal{S} \equiv \left[ \psi_1,\ldots,\psi_{\left|\mathcal{Y}^\mathcal{S}\right|} \right]^\intercal$.  Similarly, for unseen classes we obtain $A^\mathcal{U} \equiv \left[ \psi_1,\ldots,\psi_{\left|\mathcal{Y}^\mathcal{U}\right|} \right]^\intercal$. Finally, evaluation in the GZSL setting considers a test set that includes both seen and unseen classes, i.e. $\mathcal{Y}^{\mathcal{SU}} =\mathcal{Y}^{\mathcal{S}}\cup\mathcal{Y}^{\mathcal{U}}$.

\subsection{Classification pipeline}
\begin{figure}[tb]
	\centering
	\subfloat[ZSL with global semantic representations (projection after aggregation)\label{fig:architechture_global_visual}]{
		\includegraphics[scale=0.66]{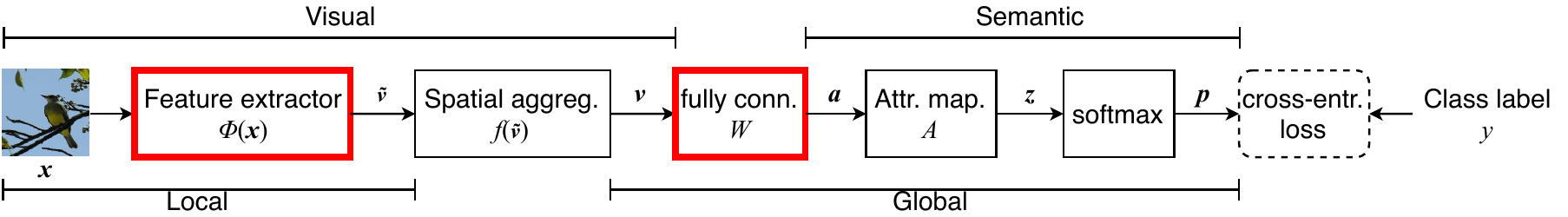}
	}\\
	\subfloat[ZSL with local semantic representations (aggregation after projection)\label{fig:architecture_local_semantic}]{
		\includegraphics[scale=0.66]{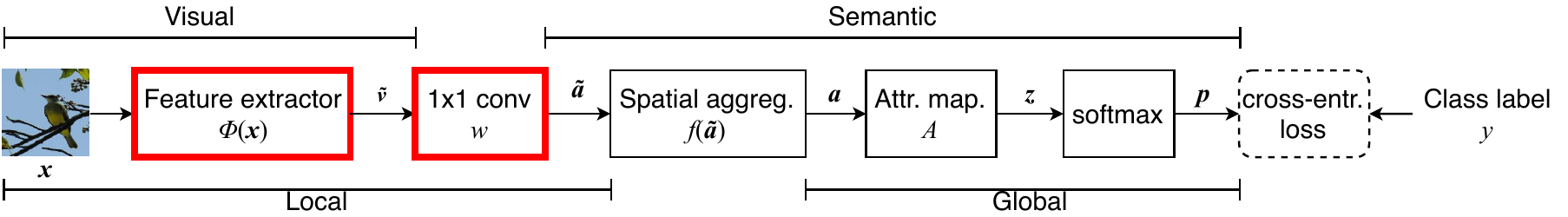}
	}\\
	\caption{Local and global representations and embedding spaces in ZSL: (a) projection to the semantic space after spatial aggregation, and (b) spatial aggregation after projection. Trainable/tunable modules are highlighted in red.}
	\label{fig:architectures}
\end{figure}

We formulate ZSL as a classification problem, using a deep convolutional neural network (CNN) that internally projects visual features to the semantic space and is trained end-to-end with cross-entropy loss on seen data (see Fig.~\ref{fig:architechture_global_visual}). In particular we are interested in some of the intermediate representations: the \textit{local visual feature} $\mathbf{\tilde{v}}\in \mathbb{R}^{M\times M\times D}$, the \textit{global visual feature} $\mathbf{v}\in \mathbb{R}^D$, the \textit{global semantic feature} $\mathbf{a}\in \mathbb{R}^L$, and the logits or \textit{unnormalized class-scores} $\mathbf{z}\in \mathbb{R}^{\left|\mathcal{Y}^\mathcal{S}\right|}$. These intermediate representations lie in three distinctive spaces: the $D$-dimensional visual space, the $L$-dimensional semantic space (where $L$ is the number of attributes in our case) and the ${\left|\mathcal{Y}^\mathcal{S}\right|}$-dimensional class space.
For convenience, we can also split the deep network into several modules: the \textit{feature extractor} $\mathbf{\tilde{v}}=\phi\left(\mathbf{x}\right)$, the \textit{spatial aggregation} operation $\mathbf{v}=f\left(\mathbf{\tilde{v}}\right)$, and the linear projection to the semantic space $\mathbf{a}=W\mathbf{v}$, parametrized by the projection matrix $W\in \mathbb{R}^{L\times D}$, i.e. a fully connected layer. The projection matrix is trainable, while the feature extractor is usually pretrained and can be optionally fine-tuned. Finally, the overall loss to minimize is
\begin{equation}
\mathcal{L}=\mathbb{E}_{\left(\mathbf{x},y\right)\sim\mathcal{S}}\left[\mathcal{CE}\left(\softmax\left(A^\mathcal{S}Wf\left(\phi\left(\mathbf{x}\right)\right)\right),y\right)\right]
\end{equation}
where $\mathcal{CE}$ is the cross-entropy loss. During test, the predicted class $y_{i^*}$ is the one with highest cross-product score
\begin{equation}
i^*=\argmax_{i} \psi^\intercal_iWf\left(\phi\left(\mathbf{x}\right)\right)
\end{equation}
with $y_i\in \mathcal{Y}^\mathcal{U}$ for ZSL, and $y_i\in \mathcal{Y}^\mathcal{SU}$ for GZSL.

A common choice for spatial aggregation is global average pooling (GAP), and a pretrained network in ImageNet as feature extractor. GoogleNet~\cite{szegedy2015going} and ResNets~\cite{he2016deep} that fit in this case, and those global features for common ZSL datasets are commonly provided off-the-shelf and used in benchmarks~\cite{Xian2018ZSLGBU}. Thus, we can also conveniently compare to those previous methods using GAP and fixing the feature extractor. In addition, this pipeline enjoys several advantages: is easy to train, has few additional parameters (i.e. the projection matrix $W$) and no additional hyperparameters. This simplicity allows our method to generalize better to unseen classes, resulting in a lower bias to seen classes (see Table~\ref{tab2}), which is critical in GZSL.

\subsection{Localized attribute representations}

Previous approaches using local representations perform localization (via part detection or attention) in the visual space, then extract local visual features and eventually aggregate them to a global visual representation, which is projected to the semantic space. We instead propose projecting local visual features to the semantic space, obtaining localized semantic representations, i.e. \textit{localized attributes} in our case. We modify our classification baseline by switching the order of spatial aggregation and projection. Now projection is performed first using a $1\times 1$ convolution with kernel $w\in \mathbb{R}^{1\times 1\times D\times L}$ resulting from reshaping $W$. Spatial aggregation is performed on the resulting localized attribute representation (see Fig.~\ref{fig:architecture_local_semantic}). The resulting global semantic representation is $\mathbf{a}=A^\mathcal{S}f\left(w \ast \phi\left(\mathbf{x}\right)\right)$.

Localized attributes provide a representation where attribute information is explicitly disentangled, where each map corresponds to a different attribute (see Fig.~\ref{fig:attribute_maps}), and every attribute has a unique attribute map. In contrast, attention maps and discovered regions essentially weight visual representations or guide the extraction of visual representations. Thus, they are not necessarily disentangling attribute information (e.g. one region could be related to many attributes or even to none), and also suffer from a more limited number of regions or attention maps, compared to the number of attributes.

In our approach, no explicit attention or detection module performs localization. In contrast, we rely on the implicit localization of visual features that the feature extractor already performs. This highlights that fine-tuning the feature extractor is often crucial to improve the discriminability and transferability of the proposed local semantic representations. 

\subsection{Spatial aggregation with localized attributes}
\label{spatial aggregation}

In this section, we investigate the role of spatial aggregation in the semantic space. While aggregating local visual representations, GAP has been proved a very effective strategy. However, localized semantic representations may behave differently, and the proposed localized attributes provide a rich and highly disentangled representation where averaging may not be the best strategy. In general, the choice of aggregation strategy is also related to how local the attributes are in a particular task. For instance, attributes in fine-grained datasets such as CUB are very local (e.g., `has\_wing\_pattern\_spotted', `has\_throat\_color\_orange'). In contrast, other datasets such as SUN contain global attributes or attributes covering wide areas (e.g.,`man-made',`trees'). We study two aggregation strategies: GAP and \textit{global maximum pooling} (GMP).

\begin{figure}[htbp]
	\centering
	\subfloat[Attribute maps (top: GAP, bottom: GMP)\label{fig:attribute_maps}]{
		\includegraphics[width=\textwidth]{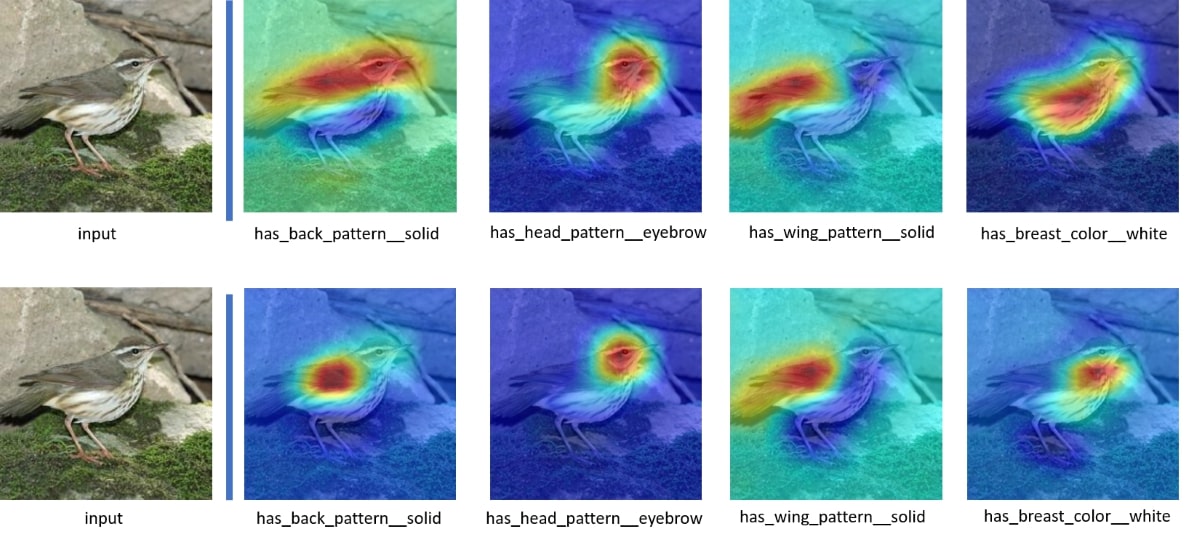}
	}\\
	\subfloat[GAP (seen)\label{fig:features_gap_seen}]{
		\includegraphics[width=0.23\textwidth]{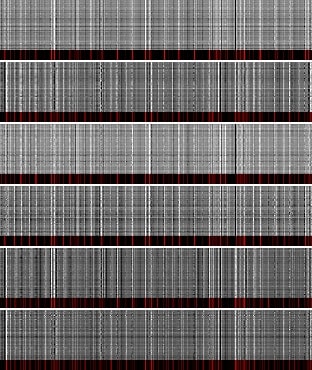}
	}
	\subfloat[GAP (unseen)\label{fig:features_gap_unseen}]{
		\includegraphics[width=0.23\textwidth]{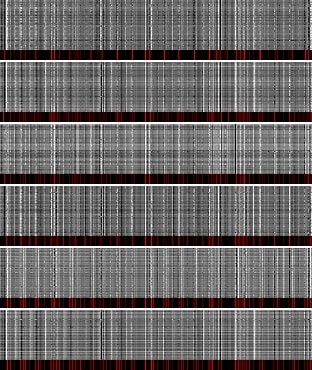}
	}
	\subfloat[GMP (seen)\label{fig:features_gmp_seen}]{
		\includegraphics[width=0.23\textwidth]{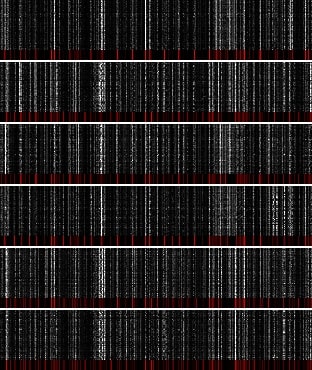}
	}
	\subfloat[GMP (unseen)\label{fig:features_gmp_unseen}]{
		\includegraphics[width=0.23\textwidth]{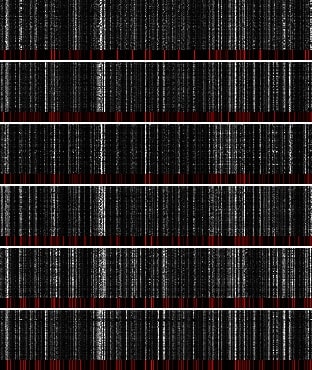}
	}\\
	\caption{Comparison between spatial aggregation methods (GAP and GMP) in CUB (fine-tuned feature extractor). Note that the attribute maps of GMP more accurately identify the relevant regions. (b-e) Global semantic representations (rows) of 50 images per class (randomly selected) of 6 classes (super-rows). Each column corresponds to one of the 312 attributes. The description corresponding to the class is shown in red. Note that GMP generates sparser and more discriminative feature patterns than GAP.}
	\label{fig:spatial_aggregation}
\end{figure}

It is worth observing that GAP is a linear operation, so local linear projection followed by GAP and GAP followed by global linear projection are equivalent. In other words, the approaches in Figs.~\ref{fig:architechture_global_visual} and \ref{fig:architecture_local_semantic} are equivalent\footnote{In our experiment, the bias term in 1x1 convolution does not influence the results.} when the spatial aggregation method is GAP. Therefore, our method with off-the-shelf pretrained feature extractors that are already using GAP, such as GoogleNet or ResNet, is already implicitly localizing attributes. When implemented as Fig.~\ref{fig:architechture_global_visual}, localized attributes are never explicitly computed. However, they can be recovered by reshaping $W$ as $w$, and computing $\mathbf{\tilde{a}}=w \ast \mathbf{\tilde{v}}$. 

Regardless of the aggregation method, our representation already achieves pretty good localization of the attributes, as shown in Fig.~\ref{fig:attribute_maps}. While GAP considers all locations equally important, GMP focuses on the most salient location of each attribute map. This can be useful in datasets such as CUB where each attribute is very localized in a single small area. In addition, when combined with fine-tuning, this encourages the feature extractor to generate maps where salient regions are smaller than with GAP (compare the effect of GAP and GMP on the attribute maps on Fig.~\ref{fig:attribute_maps}). An important difference with GAP is that GMP is not a linear operation, so the order of projection and aggregation matters in this case. Since we are interested in aggregating in the semantic space, GMP is performed after the convolution (as in Fig.~\ref{fig:architecture_local_semantic}).

The aggregated representations, i.e. the global semantic representations, obtained with our method (see Fig.~\ref{fig:spatial_aggregation}b-e) are very discriminative and robust. The patterns of the representations obtained for unseen classes are very similar to those obtained for seen classes (compare Fig.~\ref{fig:spatial_aggregation}b and c for GAP, and Fig.~\ref{fig:spatial_aggregation}d and e for GMP), which suggest that the usual bias towards seen classes is relatively low. Finally, we can compare the effect of the aggregation strategy on the global semantic feature. In particular we observe that GMP, by focusing on the most salient location for each attribute, generates sparser and arguably more discriminative feature patterns than GAP, which seems to be more sensitive to noise.

\section{Experiments}\label{sec:experiments}

{
	\renewcommand{\arraystretch}{1}
	\begin{table*}[!tb]
		\centering
		\caption{Zero-shot learning results on SUN, CUB, and AWA. PS = Proposed Split. The results report top-1 accuracy in \%. The $\dagger$ means adopting ResNet101 as feature extractor. We highlight the best result for fine-tuning and non-finetuning respectively.}
		\begin{tabular}{l  c |c |c }
			&  \multicolumn{1}{c }{\textbf{SUN}} & \multicolumn{1}{c }{\textbf{CUB}}&\multicolumn{1}{c }{\textbf{AWA2}}  \\
			\textbf{Method} & \textbf{PS}  & \textbf{PS} & \textbf{PS} \\     
			\hline
			\multicolumn{4}{c}{\textbf{Without fine-tuning}}\\
			\hline
			$\dagger$SYNC~\cite{Changpinyo2016ZSL} & $56.3$ & $55.6$ & $44.6$  \\ 
			$\dagger$DEVISE~\cite{Frome2013Devise}  & $56.5$ & $52.0$ & $59.7$   \\
			$\dagger$ALE~\cite{akata2015label} & $58.1$ & $54.9$ & $62.5$  \\
			$\dagger$PSR~\cite{annadani2018preserving} & $61.4$ & $56.0$ & $63.8$  \\
			$\dagger$DCN~\cite{liu2018generalized}  & $61.8$ & $56.2$ & $65.2$  \\
			$\dagger$MIIR~\cite{cacheux2019modeling}  & $\bm{63.5}$ & $63.8$ & $\bm{67.9}$  \\
			\hline
			\multicolumn{4}{c}{\textbf{Ours without fine-tuning}}\\
			\hline
			$\dagger$\textbf{SELAR-GAP}&$57.8$ & $57.7$ & $64.2$  \\ 
			$\dagger$\textbf{SELAR-GMP}& $58.3$ & $\bm{65.0}$ & $57.0$  \\ 
			\hline
			\hline
			\multicolumn{4}{c}{\textbf{Attention model based with fine-tuning}}\\
			\hline
			$\dagger$AREN~\cite{Xie_2019_CVPR}  & $60.6$ & $\bm{72.5}$ & $67.9$  \\ 
			$\dagger$JLA~\cite{lijoint}  &$59.6$ & $59.4$ & $\bm{70.4}$\\
			AttentionZSL~\cite{Liu_2019_ICCV} &$61.5\dagger$ & $67.6^*$ & $68.1^*$  \\
			\hline
			\multicolumn{4}{c}{\textbf{Part detection based with fine-tuning}}\\
			\hline
			SGMA~\cite{zhu2019semantic} & $-$ & $71.0^*$ & $68.8^*$  \\ 
			GAZSL~\cite{zhu2018generative} &$61.3$ & $-$ & $58.9$  \\ 
			CIZSL~\cite{elhoseiny2019creativity} & $\bm{63.7}$ & $-$ & $67.8$  \\ 
			\hline
			\multicolumn{4}{c}{\textbf{Ours with fine-tuning}}\\
			\hline
			\textbf{SELAR-GAP}$^*$&$-$ & $68.7^*$ & $64.0^*$  \\ 
			\textbf{SELAR-GMP}$^*$& $-$ & $68.1^*$ & $62.7^*$  \\ 
			$\dagger$\textbf{SELAR-GAP}&$61.4$ & $70.4$ & $66.7$  \\ 
			$\dagger$\textbf{SELAR-GMP}&$61.4$ & $63.8$ & $57.9$  \\ 
			\hline
		\end{tabular}
		\label{tab1}
	\end{table*}
}

{
	\renewcommand{\arraystretch}{1}
	\setlength{\tabcolsep}{2pt}
	\begin{table}[!tb]
		\centering
		\caption{Generalized Zero-Shot Learning on Proposed Split (PS). U = Top-1 accuracy on $\mathcal{Y}_{\mathcal{U}}$, S = Top-1 accuracy on $\mathcal{Y}_{\mathcal{S}}$, H = harmonic mean, S/U can show the bias towards seen class, $\bar{H}$ denotes the average over the H on three datasets. The underline means the second high result.  $\dagger$ indicates results using ResNet101 as feature extractor. * indicates results using VGG19 as feature extractor. We highlight the best result with fine-tuned and fixed feature extractors respectively.}
		\resizebox{\columnwidth}{!}{%
			\begin{tabular}{l  c c c c| c c c c| c c cc|c}
				&  \multicolumn{4}{c}{\textbf{SUN}} & \multicolumn{4}{c}{\textbf{CUB}} &  \multicolumn{4}{c}{\textbf{AWA2}}  &  \\
				\textbf{Method}  & \textbf{U} & \textbf{S} & \textbf{H}& \textbf{S/U}  &\textbf{U} & \textbf{S}  & \textbf{H}& \textbf{S/U} & \textbf{U} & \textbf{S} & \textbf{H}& \textbf{S/U}&$\bar{H}$\\
				\hline
				\multicolumn{13}{c}{\textbf{Without fine-tuning}}\\
				\hline
				$\dagger$SYNC~\cite{Changpinyo2016ZSL}&  $7.9$ & $\bm{43.3}$ & $13.4$ &$5.48$ & $11.5$  & $\underline{70.9}$ & $19.8$  &$6.17$& $10.0$ & $\bm{90.5}$ & $18.0$   &$9.05$&$17.1 $ \\
				$\dagger$DEVISE~\cite{Frome2013Devise}  & $16.9$ & $27.4$ & $20.9$  &$1.62$& $23.8$ & $53.0$ & $32.8$  &$2.23$& $17.1$ & $74.7$ & $27.8$   &$4.37$ &$27.2$ \\
				$\dagger$ALE~\cite{akata2015label} & $21.8$ & $33.1$ & $26.3$  &$1.52$& $23.7$ & $62.8$ & $34.4$  &$2.65$& $14.0$ & $81.8$ & $23.9$   &$5.84$&$28.2 $\\
				$\dagger$PSR~\cite{annadani2018preserving} & $20.8$ & $37.2$ & $26.7$  &$1.79$& $24.6$ & $54.3$ & $33.9$  &$2.21$& $\underline{20.7}$ & $73.8$ & $\underline{32.3}$  &$3.57$ &$31.0 $\\
				$\dagger$DCN~\cite{liu2018generalized}& $\bm{25.5}$ & $37.0$ & $\bm{30.2}$  &$1.45$& $28.4$ & $60.7$ & $38.7$  &$2.14$& $25.5$ & $84.2$ & $39.1$   &$3.30$&$36.0 $\\
				$\dagger$MIIR~\cite{cacheux2019modeling} & $22.0$ & $34.1$ & $26.7$  &$1.55$& $30.4$ & $65.8$ & $41.2$  &$2.16$& $17.6$ & $87.0$ & $28.9$   &$4.94$&$32.3 $\\
				\hline
				\multicolumn{13}{c}{\textbf{Ours without fine-tuning}}\\
				\hline
				$\dagger$\textbf{SELAR-GAP}& $\underline{23.8}$ & $32.0$ & $\underline{27.3}$  &$\bm{1.34}$& $\underline{32.1}$  & $63.0$ & $\underline{42.5}$  &$1.96$& $12.0 $ & $\underline{87.2} $ & $21.0 $  &$7.27$&$30.3 $\\
				$\dagger$\textbf{SELAR-GMP}&$22.8 $ & $31.6$ & $26.5$  &$1.39$& $\bm{43.5}$  & $\bm{71.2}$ & $\bm{54.0}$  &$\bm{1.64}$& $\bm{31.6} $ & $80.3 $ & $\bm{45.3} $   &$\bm{2.54}$&$\bm{41.9} $\\
				\hline
				\hline
				\multicolumn{13}{c}{\textbf{Attention model based with fine-tuning}}\\
				\hline
				
				$\dagger$AREN~\cite{Xie_2019_CVPR}& $19.0$ & $38.8$ & $25.5$  &$2.04$& $38.9$  & $\underline{78.7}$ & $52.1$ &$2.02$& $17.5$ & $\underline{93.2}$ & $29.5$  &$5.33$&$35.7 $ \\
				$\dagger$JLA~\cite{lijoint}& $23.2$ & $36.6$ & $28.4$  &$1.58$& $36.6$  & $59.8$ & $45.4$  &$1.63$& $24.5$ & $91.6$ & $38.3$  &$3.74$ &$37.4 $\\
				AttentionZSL~\cite{Liu_2019_ICCV}&  $18.5$ & $\bm{40.0}$ & $25.3\dagger$  &$2.16$& $36.2$  & $\bm{80.9}$ & $50.0^*$  &$2.23$& $27.0$ & $\bm{93.4}$ & $41.9^*$   &$3.46$&$39.1 $\\
				\hline
				\multicolumn{13}{c}{\textbf{Part detection based with fine-tuning}}\\
				\hline
				SGMA~\cite{zhu2019semantic}&  $-$ & $-$ & $-$ &$-$ & $36.7$  & $71.3$ & $48.5^*$ &$1.94$ & $\bm{37.6}$ & $87.1$ & $\bm{52.5^*}$  &$2.32$ &$-$\\ 
				GAZSL~\cite{zhu2018generative}&  $-$ & $-$ & $26.7$ &$-$ & $-$  & $-$ & $-$ &$-$ & $-$ & $-$ & $15.4$  &$-$ &$-$\\ 
				CIZSL~\cite{elhoseiny2019creativity}&  $-$ & $-$ & $27.8$ &$-$ & $-$  & $-$ & $-$ &$-$ & $-$ & $-$ & $24.6$  &$-$&$-$ \\ 
				\hline
				\multicolumn{13}{c}{\textbf{Ours with fine-tuning}}\\
				\hline
				$^*$\textbf{SELAR-GAP}& $-$ & $-$ & $-$  &$-$& $37.1$  & $73.2$ & $49.2^*$  &$1.97$& $14.6$ & $77.0$ & $24.5^*$   &$5.27$&$-$\\
				$^*$\textbf{SELAR-GMP}&$-$ & $-$ & $-$  &$-$& $\bm{51.4}$  & $75.2$ & $\bm{61.0^*}$ &$\bm{1.46}$ & $29.5$ & $80.2$ & $43.2^*$   &$2.72$&$-$\\
				$\dagger$\textbf{SELAR-GAP}& $\underline{23.4}$ & $\underline{37.2}$ & $\underline{28.7}$  &$1.59$& $39.0$  & $74.2$ & $51.1$  &$1.90$& $13.7 $ & $90.4 $ & $23.8 $  &$6.60$ &$34.5$\\
				$\dagger$\textbf{SELAR-GMP}&$\bm{23.8}$ & $\underline{37.2}$ & $\bm{29.0}$  &$\bm{1.56}$& $\underline{43.0}$  & $76.3$ & $\underline{55.0}$  &$1.77$& $\underline{32.9} $ & $78.7 $ & $\underline{46.4} $  &$\bm{2.39}$ &$\bm{43.5} $\\
				\hline
			\end{tabular} 
		}
		\label{tab2}
	\end{table}
}

\subsection{Datasets and Implementation Details}

We evaluate our method on three datasets: the fine-grained dataset CUB~\cite{WelinderEtal2010}, SUN~\cite{patterson2014sun} and AWA2~\cite{Xian2018ZSLGBU}. Among them, CUB has 11,788 images with 200 different classes of birds annotated with 312 attributes. SUN contains 14,340 images from 717 types of scenes with 102 attributes. Finally, AWA2 is a dataset with 50 categories of animals, which is composed of 37,322 images and 85 attributes. We follow the proposed split from~\cite{Xian2018ZSLGBU} which is commonly used in ZSL/GZSL, resulting in a 150/50, 645/72 and 40/10 (seen/unseen) category division for CUB, SUN and AWA2 datasets respectively.

We provide results for both the conventional zero-shot learning (ZSL) and generalized zero-shot learning setting (GZSL), but mainly focus on GZSL, the most challenging setting. We denote the accuracy on unseen classes and seen classes as $Acc_U$ and $Acc_S$, respectively, and the evaluation metric for GZSL is the harmonic mean on the accuracy of seen classes and unseen classes, calculated as $H=2*Acc_U*Acc_S/(Acc_U+Acc_S)$. We apply L2-normalization on the attribute matrix, commonly used in previous works.

We  report results with Imagenet pretrained ResNet101~\cite{he2016deep} and VGG19~\cite{simonyan2014very} as feature extractors, depending on the experiment, for fair comparison with previous methods. The size of the input image is 224$\times$224 pixels. Similarly, we report results with fixed and fine-tuned feature extractors, depending on the experiment.  As for learning rate, we use the following setting when using VGG19: for CUB dataset, the learning rates for feature extractor (FE) and convolutional layer (conv\_1x1) are 1e-3 and 0.2; the learning rate decays by a factor 0.1 after 15 epochs. For AWA2, the learning rates are 1e-5 and 0.5 for FE and conv\_1x1. When using ResNet101 on CUB and AWA2, all learning rates are 10 times smaller. For SUN, we use ResNet101 as other methods did, the learning rates are set to 1e-3 and 1e-2 for FE and conv\_1x1 respectively, and they decay by 0.1 every 6 epochs. The learning rate will just get decayed once.

\subsection{Quantitative Results for zero-shot Learning}
In the experiments, we compare our method with the most related state-of-the-art approaches, some of which design extra modules to find regions of interest or attention maps. We will show that in zero-shot learning, there is no necessity to add these extra modules. We refer to the pipeline with GAP as SELAR-GAP, and with GMP as SELAR-GMP. We compare essentially with methods using fixed feature extractors and global representations, and methods using fine-tuned feature extractors which is typically used for localized representations. In this case, we distinguish between methods with attention models and methods with part detectors. 
We do not compare with generative methods~\cite{felix2018multi,Huang_2019_CVPR,xian2018feature,xian2019f} that use a GAN or VAE to generate synthetic visual feature vectors for the unseen classes. These methods obtain excellent results, however they require access to the attribute vectors for the unseen classes during training, and not only during inference as for our method. Furthermore, these methods can be seen as a way of data augmentation, and can potentially be combined with the method we propose in this paper. 

\minisection{Fixed feature extractor.} We first evaluate our approach with a fixed ResNet-101 feature extractor, which is the most common representation in approaches using global representaitons. We report the ZSL and GZSL results in Table~\ref{tab1} and Table~\ref{tab2}, respectively. Interestingly, even without fine-tuning, SELAR-GAP can achieve very competitive performance, in particular for GZSL. Replacing the GAP with GMP, i.e. SELAR-GMP, can achieve state-of-the-art performance on CUB and AWA2. We argue that the main reason could be that the local features from ImageNet-pretrained networks generalize well to other datasets, and have very good generic localization ability. SELAR in this case simply trains a linear mapping that is enough to obtain an effective localized representation in the attribute space. We show additional attribute maps for both the fine-tuning and non-finetuning pipeline in the supplementary material.

\minisection{Fine-tuned feature extractor.}  We also evaluate performance with fine-tuned feature extractors in order to compare with methods using attention and part detection. Table~\ref{tab1} shows results for ZSL, where our two pipelines can get comparable performance on CUB and SUN datasets.

Table~\ref{tab2} shows the results for GZSL. SELAR-GMP achieves state-of-the-art performance on both CUB and SUN datasets, and get compelling results on AWA2 dataset, especially it surpasses other significantly more complex methods on the CUB dataset. Among the methods in Table~\ref{tab2}, AREN~\cite{Xie_2019_CVPR}, JLA~\cite{lijoint} and AttentionZSL~\cite{Liu_2019_ICCV} utilize extra modules to generate location or attribute attention, and SGMA~\cite{zhu2019semantic}, GAZSL~\cite{zhu2018generative} and CIZSL~\cite{elhoseiny2019creativity} have additional part detection modules. However, these methods obtain inferior results on most datasets compared to our method. SGMA achieves state-of-the-art performance on AWA2, however, it requires four forward passes through the feature extractor and a two times larger input resolution (448$\times$448 pixels). Specifically for SUN, using GAP or GMP does not make much difference for the results. We posit that this is due to the fact that attributes in SUN dataset are not always clearly localizable  (like the attribute 'natural light' and 'trees'); whether to consider all these regions (i.e. GAP) or only a single location (i.e. GMP) does not make a very significant difference. We also report the average H over all these datasets (we do not include here methods with results only on two datasets) in Table~\ref{tab2}, and our SELAR-GMP has the highest value with/without fine-tuning. Given the simplicity of our approach, we think that the excellent results of our method are rather astonishing, especially considering the often much more complicated architectures used by the compared methods.


From the results of ZSL and GZSL, we can see that our pipeline does not stand out in ZSL, but almost surpasses all other methods on the three dataset in GZSL. In GZSL, the ideal model should learn a better visual-semantic mapping during training, and this knowledge should be transfered to unseen classes. The results on CUB and AWA2 of our pipeline show that GMP for spatial aggregation indeed performs better than GAP, as discussed in Section~\ref{spatial aggregation}.

\begin{table}[!tb]
	\centering
	\caption{Ablation study of pooling operations and pooling spaces. The results are reported on the CUB dataset.
	}
	\begin{tabular}{cc|ccc}
		\multicolumn{2}{c|}{Aggregation} & \multicolumn{3}{c}{CUB}  \\
		Type   & Space & U    & S    & H           \\
		\hline
		GAP       & visual, attribute, class (all equivalent)         & $37.1$ & $73.2$ & $49.2$        \\
		GMP        & visual         & $39.1$ & $ \bm{80.7}$ & $52.7$        \\
		GMP & attribute (ours)       & $ \bm{51.4}$ & $75.2$ & $ \bm{61.0} $       \\
		GMP &  class         & $26.3 $ & $74.7 $ & $38.9  $       \\
		\hline
	\end{tabular}
	\label{table:localization}
\end{table}

\minisection{Aggregation method and aggregation space.} We investigate the optimal location to perform the aggregation of local features to global ones. The ablation study shown in Table~\ref{table:localization} evaluates GAP and GMP in three different spaces: visual, attribute and class ($\mathbf{\tilde{v}}$, $\mathbf{\tilde{a}}$ and $\mathbf{\tilde{z}}$ following Fig.~\ref{fig:architecture_local_semantic}), which also correspond to the order in which features are mapped to the different spaces in our classification pipeline.  

Since GAP is a linear mapping, and also the other operations (linear layer or 1x1 convolution, and attribute mapping), in this case pooling in either of the three spaces is equivalent. This is not the case in GMP, which is non-linear. In this case, Table~\ref{table:localization} shows that optimal location is  the attribute space (after the 1x1 convolutional layer). 

\minisection{Seen/unseen classes bias.} Since no images from unseen classes are observed during training, the network will inevitably be biased towards the seen classes. This bias may also increase when also fine-tuning the feature extractor.

We can evaluate the bias towards seen classes by comparing the accuracy in ZSL (see Table~\ref{tab1}) and the accuracy on unseen classes and also the harmonic mean H in GZSL (see Table~\ref{tab2}).  Our two variants  have close or slightly lower accuracy under ZSL, but achieve much higher accuracies on unseen and harmonic means than almost all the other methods. This shows our method is less biased towards the seen classes.

Another useful metric to compare seen-unseen biases between different methods is the ratio $S/U$ in GZSL (see Table~\ref{tab2}). A large ratio indicates large bias towards seen classes. SELAR-GMP obtains the lowest ratio on all datasets except for SUN (without fine-tuning) where SELAR-GAP obtains slightly better results. Again, we conjecture that there are two reasons behind this: our method is simple and does not add new hyperparameters making the localized attributes representation generalized well to unseen classes, and secondly,  GMP is less sensitive to noise and encourages more localized attributes.

\subsection{Visualization}

\begin{figure}[htb]\centering
	\includegraphics[width=1\textwidth]{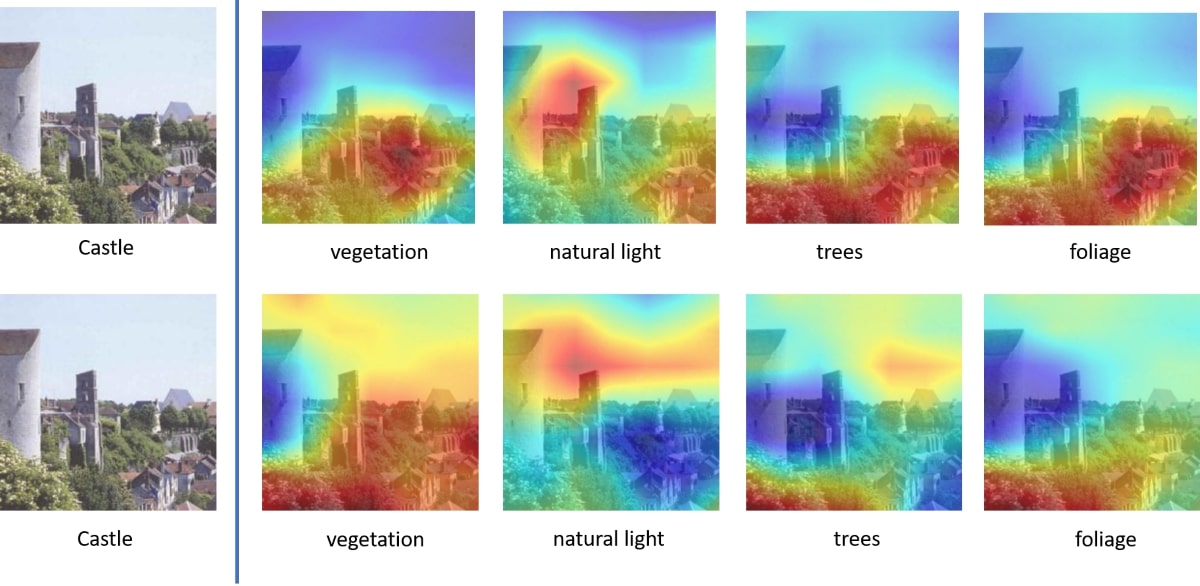}
	\caption{Visualization of attribute maps on SUN dataset from the SELAR-GAP (top one) and SELAR-GMP (bottom one). Below the image are the corresponding attributes.}
	\label{am_sun}
\end{figure}

\minisection{Localized attribute map on SUN dataset.} Since we have shown the visualization of attribute maps on CUB in Section~\ref{spatial aggregation}, here we visualize some attribute maps $\tilde{a}$ in the localized attribute space, for both SELAR-GAP and SELAR-GMP (with fine-tuning) on SUN dataset in Fig.~\ref{am_sun}. Whereas on the CUB and AWA datasets the attributes are present in clearly localizable small regions, for the SUN dataset this is not the case. This maybe the reason why the SELAR-GAP and SELAR-GMP have similar performance on SUN.

\begin{figure}[htb]\centering
	\includegraphics[width=1\textwidth]{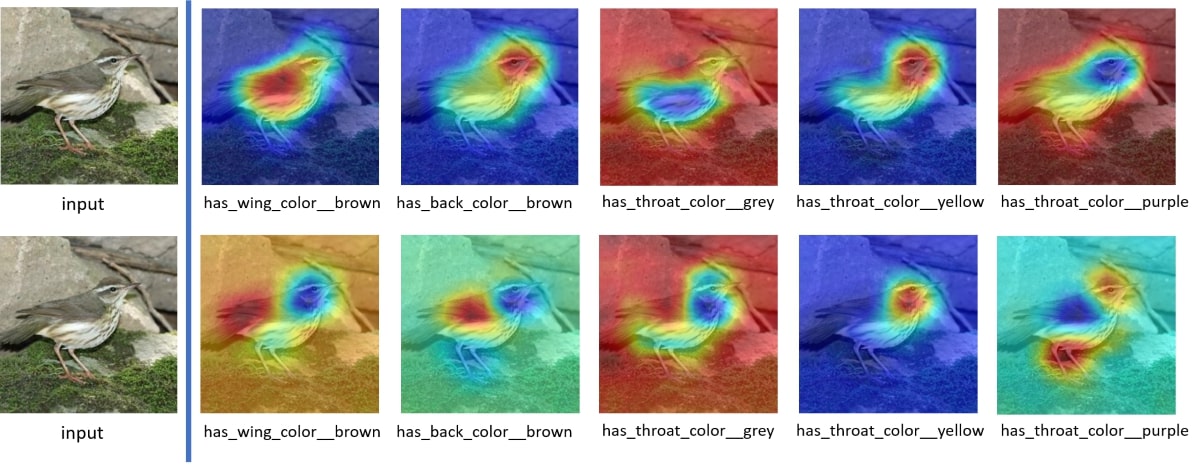}
	\caption{Visualization of attribute maps on CUB dataset from the SELAR-GAP (upper part) and SELAR-GMP (lower part). Below the image are the corresponding attributes. Those attributes has \textbf{lower} value (decreasing from left to right) in the attribute vector.}
	\label{middle_low}
\end{figure}

One thing to emphasize is that we can not guarantee that each attribute map really corresponds to the true attribute, since there is no other constraint. For example, the attribute map for attribute 'neck color - red' is not necessarily localizing the neck region, the network learns to correlate the related region with its attribute value automatically. But we find that those feature maps with high attribute values ($>$85) in the attribute vector are always highly related to the corresponding attribute. We show additional attribute maps with lower attribute values in Fig.~\ref{middle_low}, those attributes has either middle value or do not exist (0) in the attribute vector of that class, you can see sometimes the attribute map indeed corresponds to the specific attribute, but sometimes not.

\section{Conclusions}
In this paper, we focus on localized semantic representations, and provide a simple but effective pipeline for zero-shot learning, dubbed as SELAR. In this pipeline the localized attribute can be obtained. Each feature map in the localized attribute space corresponds to one specific attribute. We also study the role of spatial aggregation to improve the localization ability in the localized attribute space, and show that global max pooling can lead to significant performance improvement in generalized zero shot learning. This is mainly caused by a drastic improvement on the unseen classes.  Finally, we achieve state-of-the-art performance on CUB and AWA2 dataset under both fine-tuning and non-finetuning setting, and also obtain compelling results on AWA2 dataset. This simple pipeline can be a new baseline in zero-shot learning.

\section*{Acknowledgements}
We acknowledge the support from Huawei Kirin Solution, the European Union’s H2020 research under the Marie Sklodowska-Curie grant agreement No.665919 and the Spanish Government funding for projects PID2019-104174GB-I00 and RTI2018-102285-A-I00.

	%
	%
	\bibliographystyle{splncs04}
	\bibliography{egbib}

\end{document}